\documentclass[runningheads]{llncs}
\usepackage[T1]{fontenc}
\usepackage{graphicx}
\usepackage{amsmath}
\usepackage{multirow}
\usepackage{xcolor}
\usepackage{amssymb}
\usepackage{graphicx}
\usepackage[misc]{ifsym}
\begin{document}
\title{ViSMoE: Visual-Aware Sparse Mixture-of-Experts for Embodied Referring Expression Grounding}
%
%
%
\author{Shuo Feng, Piji Li\textsuperscript{\Letter}
}
\institute{College of  Artificial Intelligence, \\
Nanjing University of Aeronautics and Astronautics, Nanjing, 211106, China\\
MIIT Key Laboratory of Pattern Analysis and Machine Intelligence, Nanjing, 211106\\
The Key Laboratory of Brain-Machine Intelligence Technology, \\Ministry of Education, Nanjing, 211106, China.\\
\email{\{fengshuo, pjli\}@nuaa.edu.cn}}

\maketitle              
\begin{abstract}
Embodied Referring Expression Grounding is the task of enabling an agent to navigate in real environments and to localize a remote object based on natural language instructions. In this scenario, the agent needs to select one view for navigation at each step and identify a specific object among all candidate objects at the destination. However, most of the previous approaches fail to distinguish between views and objects, instead processing them using the vanilla vision encoder, which results in ambiguous representations of both views and objects. To address the above issues, we propose ViSMoE, which equips sparse Mixture-of-Experts with a visual-aware routing policy for the embodied agent. This framework processes different types of visual information specifically, resulting in discriminative visual representations for both views and objects. Experimental results on REVERIE and SOON datasets demonstrate that ViSMoE outperforms the previous state-of-the-art methods, showing the superiority of our proposed method.

\end{abstract}
\section{Introduction}

Developing robots that can execute tasks based on natural language instructions has been a longstanding objective in AI and robotics. In this field, Vision-and-Language Navigation (VLN)~\cite{anderson2018vision} has recently attracted increasing attention from many researchers due to its broad real-world applications. 

In VLN, navigation instructions are mainly divided into two types, \textit{i.e.}, step-by-step instructions such as R2R \cite{anderson2018vision} and goal-oriented instructions such as REVERIE \cite{qi2020reverie}. 
Compared to detailed step-by-step instructions, high-level goal-oriented instructions are more prevalent in practical applications and present greater technological challenges. In this paper, we address the task of Embodied Referring Expression Grounding, specifically focusing on REVERIE \cite{qi2020reverie} and SOON \cite{zhu2021soon}. In this task, the agent selects a discrete view to determine the next navigation step and identifies the target object among all candidate objects present at the destination, based on the provided instruction. Therefore, it is crucial to establish a comprehensive framework that effectively models both views and objects, addressing the dual challenges of language-guided navigation and referring expression grounding within Embodied Referring Expression Grounding.

\begin{figure}[htbp]
    \centering
    \hspace{-3mm}
    \includegraphics[width=0.95\textwidth]{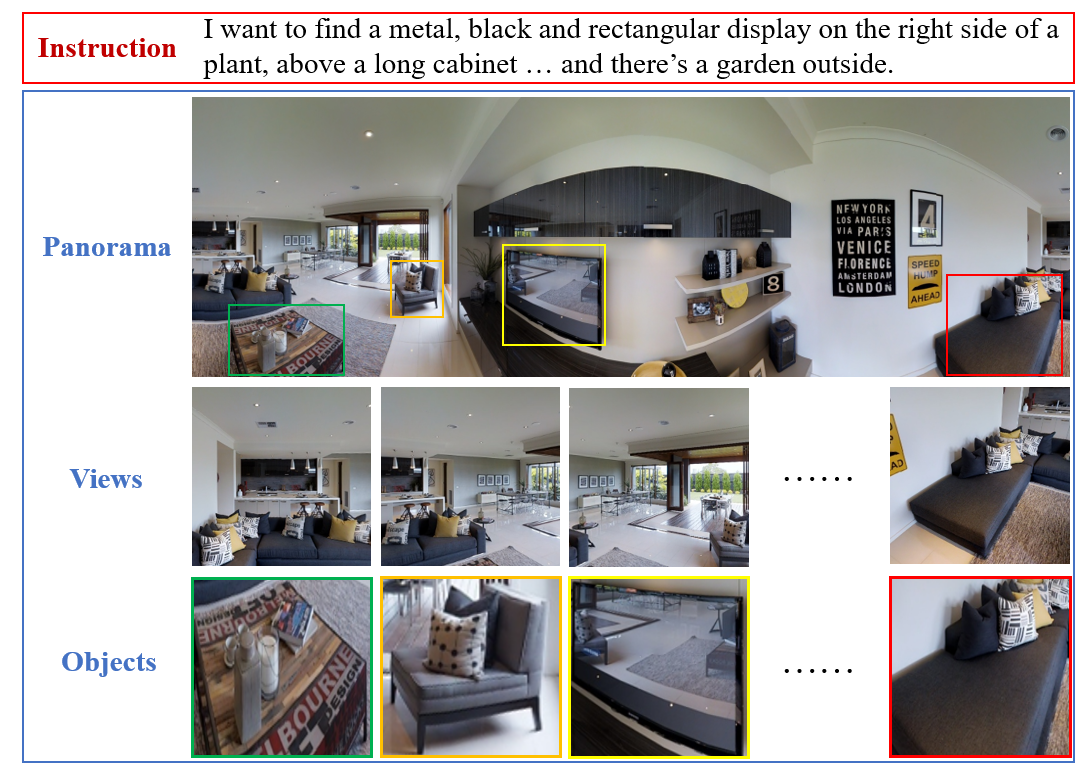} 
    \caption{Illustration of navigation instruction, panorama, discrete views and candidate objects in SOON dataset. Discrete views provide a depiction of the scene layout, while candidate objects offer detailed information about specific items.}
    \vspace{-3mm}
    \label{fig:example}
\end{figure}
Previous approaches \cite{chen2022think,wang2023gridmm} treat discrete views and candidate objects indistinguishably by concatenating their respective embeddings to form a unified visual input. The visual input is then fed directly into the panorama encoder \cite{chen2021history} and cross-model encoder to derive the contextual representations. Nevertheless, as shown in Figure \ref{fig:example}, though both discrete views and candidate objects belong to the visual input, there exist fundamental distinctions between them. Discrete views offer comprehensive overviews of the room layout, capturing the spatial relationships and broader environmental context, whereas candidate objects focus narrowly on detailed information about specific items within the scene. Moreover, discrete views are selected sequentially for the subtask of language-guided navigation, while candidate objects are identified upon reaching the destination for the subtask of referring expression grounding. In brief, discrete views and candidate objects not only exhibit differences in their data distribution but are also tied to distinct tasks, reflecting a multi-input and multi-task challenge. Therefore, directly employing the vanilla encoder to process both types of inputs simultaneously introduces significant challenges.

To address the aforementioned issues, we propose a \textbf{Vi}sual-aware \textbf{S}parse \textbf{M}ixture-\textbf{o}f-\textbf{E}xperts (ViSMoE) framework, which integrates sparse Mixture-of-Experts (SMoE) \cite{jacobs1991adaptive} with the visual-aware routing policy. The ViSMoE incorporates multiple experts within its architecture, where each expert is specialized for a specific task or feature subspace, thus effectively addressing the challenges of handling multi-input and multi-task. In addition, we introduce the visual-aware routing policy to further enhance the specialization of experts for managing diverse visual inputs. Specifically, we employ learnable type embeddings for both discrete views and candidate objects. These type embeddings are added to the embeddings of views and objects to serve as the input for the gated network, which enables our proposed ViSMoE to consider the distributional difference between views and objects, capturing more distribution-specific information.

We summarize our key contributions as follows:
\begin{itemize}
\vspace{-2mm}
\item We propose ViSMoE, which disentangles diverse task-specific visual features within the SMoE layer, tackling the challenges presented by heterogeneous inputs.
\item We further introduce the visual-aware routing policy to enhance the specialization of experts in processing diverse visual inputs.
\item Extensive results on REVERIE and SOON demonstrate the effectiveness of our framework, highlighting the benefits of using MoE for processing visual features in embodied AI.
\end{itemize}

\section{Related Work}
\noindent\textbf{Embodied Referring Expression Grounding} requires the agent to find an optimal path to the target by sequentially selecting the next viewpoint and to identify a specific object at the destination. 
It is critical to model the current observations as they are directly tied to navigation decisions and object grounding. One line of work \cite{li2023kerm,mohammadi2024augmented} focuses on incorporating knowledge into view representation to enhance vision-language alignment. 
Additionally, \cite{mohammadi2024augmented} leverages corresponding object features to enrich the semantics of view representations. 
Previous methods focus on enhancing visual representations for views through new representation fusion techniques, with limited attention to object modeling. In contrast, our work introduces a novel Visual-Aware SMoE to better capture distribution-specific features for both views and objects, treating them with equal importance.\\
\textbf{Mixture-of-Experts (MoE)} has been widely explored across various domains \cite{jordan1994hierarchical,yuksel2012twenty}, achieving significant advancements due to its divide-and-conquer strategy, which employs a series of specialization experts. Subsequent works \cite{shazeer2017outrageously} propose a sparsely activated method for utilizing experts, leading to the development of sparse MoE (SMoE), which reduces the cost at the training and inference stage while maintaining effectiveness. Several recent studies have applied SMoE in multi-task scenarios. Specifically, \cite{gupta2022sparsely,chen2023adamv} utilize task-specific router networks to activate relevant model components for each task, achieving excellent experimental results on multiple NLU tasks \cite{wang2018glue} and visual dense prediction tasks \cite{lin2014microsoft}. Besides, SMoE has demonstrated significant success in addressing the distributional difference in multi-input scenarios. For instance, EVE \cite{chen2024eve} utilize a shared transformer to encode the text and image input. Compared to them, the unified task of navigation and object grounding in Embodied Referring Expression Grounding satisfies both the multi-input and multi-task settings. We tackle this unified task using a method called ViSMoE, which equips the SMoE with the visual-aware router.

\section{Method}
\subsection{Task Formulation}

In Embodied Referring Expression Grounding \cite{qi2020reverie,zhu2021soon}, given a high-level natural language instruction $ \mathcal{W} =\{w_1, w_2, \cdots, w_L \}$, where $w_i$ denotes the $i^{th}$ token and $L$ is the length of the instruction, the agent needs to navigate through an indoor environment and localize a remote object at the destination. The indoor environment is represented as an undirected connectivity graph $\mathcal{G} = (\mathcal{V}, \mathcal{E})$, where $\mathcal{V}$ represents the navigable nodes and $\mathcal{E}$ indicates the connectivity between them. At the time step $t$, the agent perceives a panoramic view $\mathcal{V}_t=\{v_{t,i}\}_{i=1}^{n}$.
The agent decides an action $a_t$ from a special ``stop'' token and the navigable viewpoints $\mathcal{N}(\mathcal{V}_t)=\{v_{t,i}\}_{i=1}^{K}$, where $\mathcal{N}(\mathcal{V}_t) \subseteq \mathcal{V}_t$. To enhance detailed visual perception and mitigate the challenges of direct object detection from the panorama \cite{chen2022think}, candidate object features $\mathcal{O}_t=\{o_{t,i}\}_{i=1}^{m}$ are extracted using annotated object bounding boxes \cite{qi2020reverie} or object detectors. When the agent decides to stop at a location, it needs to select an object from the candidate objects.


\subsection{Base Model}
Inspired by \cite{chen2022think}, we design our model with the following key components, which employs dual-scale reasoning by leveraging both local and global observations.\\
\textbf{Language Encoder.} Each word embedding $w_i$ is added with its corresponding position embedding in the sentence. The resulting embeddings are then fed into a multi-layer transformer \cite{vaswani2017attention} to get the contextual language representations $\hat{\mathcal{W}}$.\\
\textbf{Vision Encoder.} Discrete views $\mathcal{V}_t$ and candidate objects $\mathcal{O}_t$ are added their corresponding relative angle features $a_t$, bounding box features $b_t$ and navigation type embeddings $n_t$, which can be described as:
\begin{align}
    \mathcal{V}_t' = \text{LN}(\text{W}_1^{\mathcal{V}}\mathcal{V}_t)+\text{LN}(\text{W}_2[a_t^{\mathcal{V}};b_t^{\mathcal{V}}])+n_t^{\mathcal{V}}
    \tag{1}
    \label{image} 
\end{align}
where the $\text{LN}$ denotes layer normalization, $\text{W}_1^{\mathcal{V}}$ and $\text{W}_2$ are learnable parameters, we can derive the expression for $\mathcal{O}_t'$ by replacing $\mathcal{V}$ with $\mathcal{O}$ in Equation~\ref{image}, $n_t^{\mathcal{V}}$ incudes two type of embeddings: one for candidate views and another for non-candidate view. Similarly, $n_t^{\mathcal{O}}$ represents the type embedding for objects. A special “stop” token $\mathcal{V}_{t,0}$ is added to $\mathcal{V}_{t}$ for the stop action. They are then fed into a panorama encoder (\textit{i.e.} a two-layer transformer) to obtain the contextual vision representations $\mathcal{V}_t^c$ and $\mathcal{O}_t^c$:
\begin{align}
    [\mathcal{V}_t^c;\mathcal{O}_t^c] = \text{Pano}([\mathcal{V}_t';\mathcal{O}_t'])
    \tag{2}
    \label{pano} 
\end{align}
\textbf{Dual-Scale Cross-Modal Encoder.} The topological graph $\mathcal{G}_t = \{\boldsymbol v,\boldsymbol e \mid \boldsymbol v \subseteq \mathcal{V}, \boldsymbol e \subseteq \mathcal{E} \}$ at time step $t$ contains three types of nodes: current node, visited nodes, and navigable nodes. Current node representation $\mathcal{H}_t$ is updated by averaging the local features $\mathcal{V}_t^c$ and $\mathcal{O}_t^c$. The features of visited nodes remain unchanged. Navigable nodes, which may have been partially observed from different viewpoints in the past trajectory, are represented by averaging all the partially observed views. We employ the cross-modal transformer to model the global-scale and local-scale cross-modal encoders, ensuring effective multi-level observation representation. For local-scale reasoning, local visual features $\mathcal{V}_t$ and $\mathcal{O}_t$ perform cross-attention over language representations $\hat{\mathcal{W}}$, resulting in $\hat{\mathcal{V}}_t$ and $\hat{\mathcal{O}}_t$:
\begin{align}
    [\hat{\mathcal{V}}_t^c;\hat{\mathcal{O}}_t^c] = \text{Cross-Attn}([\mathcal{V}_t^c;\mathcal{O}_t^c], \hat{\mathcal{W}})
    \tag{3}
    \label{cros} 
\end{align}
For global-scale reasoning, node representations $\mathcal{H}$ perform cross-attention over language representations $\hat{\mathcal{W}}$, yielding $\hat{\mathcal{H}}$:
\begin{align}
    \hat{\mathcal{H}} = \text{Cross-Attn}(\mathcal{H}, \hat{\mathcal{W}})
    \tag{4}
    \label{cros1} 
\end{align}
Global features $\hat{\mathcal{H}}$ and local features $\hat{\mathcal{V}}_t^c$ are employed for navigation. Object features $\hat{\mathcal{O}}_t^c$ are employed for object grounding.

\begin{figure*}[!t]
    \centering
    \includegraphics[width=1\textwidth]{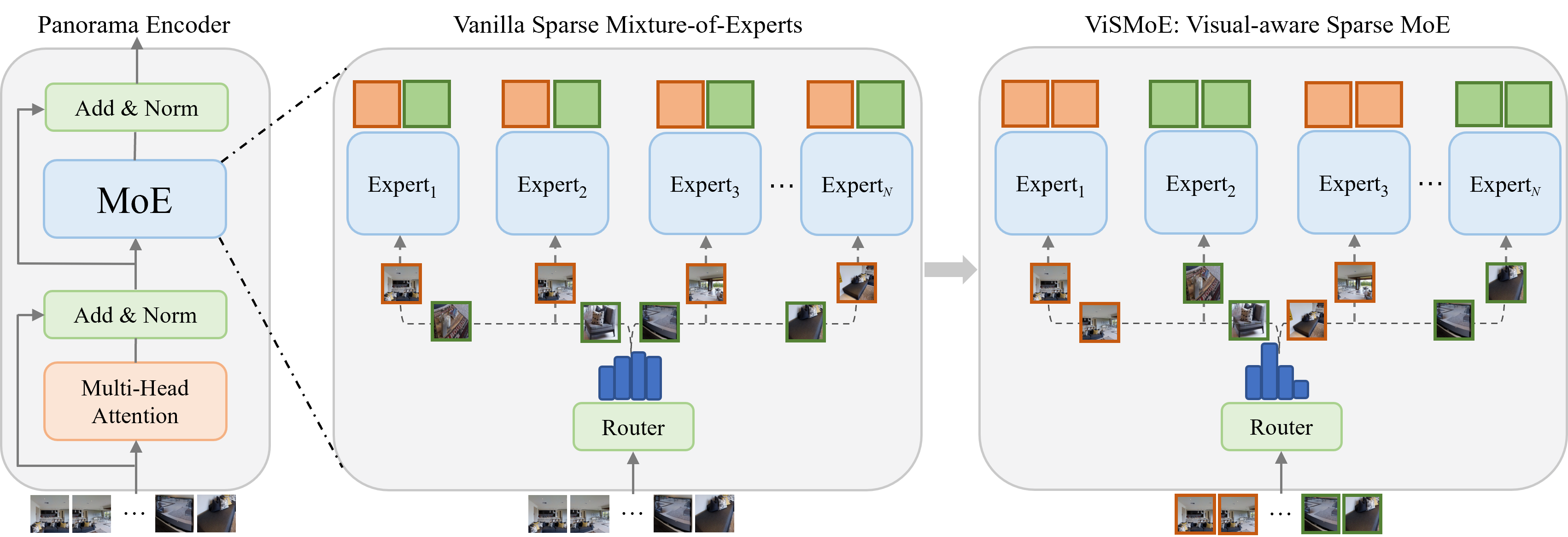} 
    \caption{Overview of ViSMoE: The dense FFN in the vision encoder is replaced by SMoE layer. We use brown boxes to denote scene images and green boxes to represent object images. Unlike vanilla SMoE, which ignores image categories, ViSMoE’s routing input includes them, allowing the trained router to select the appropriate expert based on both the input token and its visual type. Therefore, it can be seen that while the vanilla MoE router balances the distribution of input tokens across experts, ViSMoE learns to route visual tokens to specific experts (e.g., expert-1 and expert-3 for view images, expert-2 and expert-N for object images). 
    \vspace{-3mm}
    }
    \label{fig:moe-architecture}
\end{figure*}

\subsection{Visual-Aware Sparse Mixture-of-Experts}
In Embodied Referring Expression Grounding, navigation and object grounding rely on distinct visual input (\textit{i.e.} views and objects) with different data distributions: views offer comprehensive room layouts, while objects focus on specific item details. Directly using the same Feed-Forward Networks (FFNs) for all visual tokens results in coarse visual representations, which negatively impact model performance. To solve this problem, we propose a Visual-Aware Sparse Mixture-of-Experts (ViSMoE) as shown in Figure \ref{fig:moe-architecture}, which incorporates the visual-aware routing policy on the SMoE to capture distribution-specific information between views and objects. \\
\textbf{Sparse Mixture-of-Experts (SMoE).} An SMoE layer contains a gated network (router) and $N$ experts which are FFNs. Given an input token $x \in \mathcal{R}^D$, which is replaced by visual representations $\mathcal{V}$ and $\mathcal{O}$ in practice, it is fed into the router $\mathcal{G}(\cdot)$ to get the assignment probability for $N$ experts, which are used for further processing. This can be described as:
\begin{align*}
    \mathcal{G}(x) = \text{top-k}(\text{softmax}(\text{W}_g \cdot x))
    \tag{5}
\end{align*}
where $\text{W}_g \in \mathcal{R}^{N \times D}$ is learnable and top-k(·) is a function that selects the largest $k$ values.

The final output of the SMoE is the weighted sum of the k features derived from the expert selected by the router, which can be formulated as:
\begin{align*}
    \hat{x} = \sum_{i=1}^{k} \mathcal{G}(x)_i \cdot \text{FFN}_i(x)
    \tag{6}
    \label{moe}
\end{align*}
\textbf{Visual-Aware Routing.} The conditional sparse MoE imposes no specific constraints on the router network, resulting in insufficiently targeted routing for different token types. As shown in Figure \ref{fig:moe-architecture}, the vanilla SMoE distributes visual tokens of various types across all experts, whereas ViSMoE directs them to specific experts. 

Specifically, we propose a visual-aware routing policy that explicitly provides the router network with visual type information by incorporating visual type-specific embeddings. This visual-aware routing can be described as follows:
\begin{align*}
    \mathcal{G}(x) = \text{top-k}(\text{softmax}(\text{W}_g \cdot (x+b_t)))
    \tag{7}
    \label{router}
\end{align*}
where $b_t \in \mathcal{R}^D$ represents the type embedding, with $b_v$ for view representations $\mathcal{V}$ and $b_o$ for object representations $\mathcal{O}$.\\
\textbf{Auxiliary Loss.} To prevent imbalanced token routing, we introduce a Load-Balancing Loss following \cite{shazeer2017outrageously} as an auxiliary loss for the router network. It is formulated as follows:
\begin{align*}
    \mathcal{L}_{\text{Aux}} = \alpha \cdot N \sum_{i=1}^{N}p_i \times f_i
    \tag{8}
\end{align*}
where $\alpha$ is set to 0.001, $N$ is the number of experts, $p_i$ is the average routing weight for the $i^{th}$ expert, and $f_i$ is the fraction of tokens allocated to the $i^{th}$ expert. This loss function promotes an even distribution of tokens across experts.

\subsection{Training and Inference}
After incorporating Equation \ref{router}, we employ Equation \ref{moe} to replace the FFN layers in Equations \ref{pano} and \ref{cros}, enabling a more specialized and efficient processing framework. ViSMoE’s parameters are updated during pre-training and fine-tuning.\\
\textbf{Pre-training.} We pre-train the model on four tasks: masked language modeling (MLM), masked region classification (MRC), single-step action prediction (SAP) \cite{chen2021history} and object grounding (OG) \cite{lin2021scene}. The SAP and OG losses are defined as follows:
\begin{align*}
    \mathcal{L}_{\text{SAP}} = \sum_{t=1}^{T}-log \, p(a^{*}_t|\mathcal{W},P_{<t})
    \tag{9}
\end{align*}
where $P_{<t}$ is a partial ground-truth demonstrations, $a^{*}_t$ is the expert action of $P_{<t}$.
\begin{align*}
    \mathcal{L}_{\text{OG}} = -log \, p(o^{*}|\mathcal{W},P_{T})
    \tag{10}
\end{align*}
where $o^{*}$ is the ground-truth object at the destination $P_{T}$. In summary, the total objective of pre-training is:
\begin{align*}
\begin{split}
    \mathcal{L}_{\text{P}} = \mathcal{L}_{\text{MLM}} + \mathcal{L}_{\text{MRC}} + \mathcal{L}_{\text{SAP}}+ \mathcal{L}_{\text{OG}} + \mathcal{L}_{\text{Aux}}
\end{split}   
\tag{11}
\end{align*}
\textbf{Fine-tuning.} For fine-tuning, in addition to SAP, we employ the DAgger algorithm \cite{ross2011reduction} to train the navigation policy, enabling the agent to navigate based on predicted probabilities during training and to correct bias through learning:
\begin{align*}
    \mathcal{L}_{\text{DA}} = \sum_{t=1}^{T}-log \, p(a^{*}_t|\mathcal{W},\hat{P}_{<t})
    \tag{12}
\end{align*}
where $\hat{P}_{<t}$ is the sampled demonstrations. The total objective of fine-tuning is:
\begin{align*}
\begin{split}
    \mathcal{L}_{\text{F}} = \mathcal{L}_{\text{SAP}}+ \mathcal{L}_{\text{DA}} + \mathcal{L}_{\text{OG}} + \mathcal{L}_{\text{Aux}}
\end{split}   
\tag{13}
\end{align*}


\section{Experiments}
\subsection{Experimental Setup}
\label{setup}
\noindent\textbf{Datasets.} We conduct experiments and evaluate our model on benchmarks REVERIE \cite{qi2020reverie} and SOON \cite{zhu2021soon}.

\textbf{REVERIE} contains high-level instructions that average 18 words in length, with navigation paths ranging from 4 to 7 steps. Each panorama is equipped with predefined object bounding boxes.

\renewcommand{\arraystretch}{1.2}
\begin{table*}
\small
\tabcolsep=0.2cm
\centering
\caption{Evaluation on REVERIE dataset.}
\resizebox{1.0\textwidth}{!}{
\begin{tabular}{c|cccc|cc|cccc|cc}
\hline 
\multirow{3}{*}{Methods} & \multicolumn{6}{c|}{{Val Unseen}} & \multicolumn{6}{c}{{Test Unseen}}\tabularnewline
\cline{2-13} \cline{3-13} \cline{4-13} \cline{5-13} \cline{6-13} \cline{7-13} \cline{8-13} \cline{9-13} \cline{10-13} \cline{11-13} \cline{12-13} \cline{13-13}
 & \multicolumn{4}{c|}{{Navigation}} & \multicolumn{2}{c|}{{Grounding}} & \multicolumn{4}{c|}{{Navigation}} & \multicolumn{2}{c}{{Grounding}}\tabularnewline
\cline{2-13} \cline{3-13} \cline{4-13} \cline{5-13} \cline{6-13} \cline{7-13} \cline{8-13} \cline{9-13} \cline{10-13} \cline{11-13} \cline{12-13} \cline{13-13}
 & {TL\textdownarrow{}} & {OSR\textuparrow{}} & {SR\textuparrow{}} & {SPL\textuparrow{}} & {RGS\textuparrow{}} & {RGSPL\textuparrow{}} & {TL\textdownarrow{}} & {OSR\textuparrow{}} & {SR\textuparrow{}} & {SPL\textuparrow{}} & {RGS\textuparrow{}} & {RGSPL\textuparrow{}}\tabularnewline
 \hline 
 {CKR\ \cite{gao2021room}} & {26.26} & {31.44} & {19.14} & {11.84} & {11.45} & {-} & {22.46} & {30.40} & {22.00} & {14.25} & {11.60} & {-}\tabularnewline
{SIA\ \cite{lin2021scene}} & {41.53} & {44.67} & {31.53} & {16.28} & {22.41} & {11.56} & {48.61} & {44.56} & {30.80} & {14.85} & {19.02} & {9.20}\tabularnewline
{HAMT\ \cite{chen2021history}} & \textbf{14.08} & {36.84} & {32.95} & {30.20} & {18.92} & {17.28} & \textbf{13.62} & {33.41} & {30.40} & {26.67} & {14.88} & {13.08}\tabularnewline
{TD-STP\ \cite{zhao2022target}} & {-} & {39.48} & {34.88} & {27.32} & {21.16} & {16.56} & {-} & {40.26} & {35.89} & {27.51} & {19.88} & {15.40}\tabularnewline
{FDA\ \cite{he2024frequency}} & {19.04} & {51.41} & {47.57} & \textbf{35.90} & {32.06} & \textbf{24.31} & {17.30} & {53.54} & {49.62} & {36.45} & {30.34} & {22.08}\tabularnewline
\hline 
{DUET\ \cite{chen2022think}} & {22.11} & {51.07} & {46.98} & {33.73} & {32.15} & {23.03} & {21.30} & {56.91} & {52.51} & {36.06} & {31.88} & {22.06}\tabularnewline 
ViSMoE (Ours) & {23.85} & \textbf{54.98} & \textbf{50.18} & {35.35} & \textbf{33.71} & {23.87} & {22.08} & \textbf{58.63} & \textbf{54.16} & \textbf{37.27} & \textbf{33.38} & \textbf{22.49}\tabularnewline
\hline 
\end{tabular}
}
\label{table1}
\end{table*}

\begin{table*}
\small
\tabcolsep=0.2cm
\centering
\caption{Evaluation on SOON dataset.}
\resizebox{1.0\textwidth}{!}{
\begin{tabular}{c|ccccc|ccccc}
\hline
    \multirow{2}{*}{Methods} & \multicolumn{5}{c|}{Val Unseen} & \multicolumn{5}{c}{Test Unseen}\\
    \cline{2-11}
    & {TL\textdownarrow{}} & {OSR\textuparrow{}} & {SR\textuparrow{}} & {SPL\textuparrow{}} & {RGSPL\textuparrow{}} & {TL\textdownarrow{}} & {OSR\textuparrow{}} & {SR\textuparrow{}} & {SPL\textuparrow{}} & {RGSPL\textuparrow{}}\\
    \hline
    GBE~\cite{zhu2021soon} & \textbf{28.96} & 28.54 & 19.52 & 13.34 & 1.16 & \textbf{27.88} & 21.45 & 12.90 & 9.23 & 0.45 \\
    KERM~\cite{li2023kerm} & 35.83 & 51.62 & 38.05 & 23.16 & 4.04 & - & - & - & - & - \\
    GridMM~\cite{wang2023gridmm} & 38.92 & \textbf{53.39} & 37.46 & 24.81 & 3.91 & 46.20 & 48.02 & 36.27 & 21.25 & 4.15 \\
    \hline
    DUET~\cite{chen2022think} & 36.20 & 50.91 & 36.28 & 22.58 & 3.75 & 41.83 & 43.00 & 33.44 & 21.42  & 4.17\\
    Ours & 36.17 & 51.06 & \textbf{38.23} & \textbf{26.13} & \textbf{4.31} & 38.41 & \textbf{48.56} & \textbf{36.69} & \textbf{21.53} & \textbf{5.20}\\
\hline
\end{tabular}
}
\label{table2}
\end{table*}

\textbf{SOON} offers detailed instructions that precisely describe the target scenes and objects. These instructions average 47 words in length, with navigation paths varying from 2 to 21 steps. 
\\
\textbf{Evaluation Metrics.} We employ standard and widely-used evaluation metrics, including trajectory length (TL), oracle success rate (OSR), success rate (SR), and success rate penalized by path length (SPL) to assess navigation performance. Additionally, remote grounding success rate (RGS) and remote grounding success rate weighted by path length (RGSPL) are used to evaluate the object grounding task. \\
\textbf{Implementation Details.} 
We employ 9, 2, 4 and 4 transformer layers in the text encoder, panorama encoder, coarse-scale cross-modal encoder and fine-scale cross-modal encoder. Except for the panorama encoder, the other encoders' parameters are initialized with the pretrained LXMERT \cite{tan2019lxmert}. 

\subsection{Main Results}
\noindent\textbf{Results on REVERIE.} Table \ref{table1} show the performance of our ViSMoE compared to previous methods on REVERIE. Our model outperforms the above model and improves almost all metrics which demonstrates the effectiveness and generalization ability of our ViSMoE. Specifically, compared to DUET\cite{chen2022think}, our model improves SR and RGS by 3.20\% and 1.56\% on validation split, and by 1.65\% and 1.50\% on test split.
\\
\textbf{Results on SOON.} Table \ref{table2} presents the results on the SOON dataset. Our proposed ViSMoE outperforms DUET across all metrics. Especially, our ViSMoE achieves a 3.55\% improvement in SPL and a 0.56\% gain in RGSPL on the validation split, as well as a 0.11\% increase in SPL and a 1.03\% gain on the test split. Notably, our ViSMoE demonstrates a significantly larger improvement in RGSPL on the SOON dataset compared to REVERIE, highlighting its superior performance on tasks with greater visual and linguistic complexity. According to the statistics, each instruction in SOON contains an average of 11.3 nouns, significantly more than the 5.31 in REVERIE. Additionally, each navigation step in SOON includes an average of 45.8 candidate objects, compared to just 3.42 in REVERIE.
\begin{table}
\small
\noindent\begin{minipage}[t]{1\columnwidth}%
\tabcolsep=0.28cm
\renewcommand{\arraystretch}{1.3}
\centering
\caption{Ablation study on the val unseen split of REVERIE.}
\hspace{-0.3cm}
\begin{tabular}{cccc|cccc}
\hline
Baseline & SMoE & Vis. & Aux. & SR\textuparrow{} & SPL\textuparrow{} & RGS\textuparrow{} & RGSPL\textuparrow{}\tabularnewline
\hline 
 $\checkmark$ & & & & 46.98 & 33.73 & 32.15 & 23.03
\tabularnewline
$\checkmark$ & $\checkmark$ & & & 48.37 & 34.09 & 32.57 & 22.94
\tabularnewline
$\checkmark$ & $\checkmark$ &  &$\checkmark$& 48.46 & 34.19 & 32.67 & 23.16
 \tabularnewline
$\checkmark$ & $\checkmark$ & $\checkmark$ &  & 49.92 & 35.01 & 33.42 & 23.74\tabularnewline
\hline 
$\checkmark$ & $\checkmark$ & $\checkmark$ &$\checkmark$& \textbf{50.18} & \textbf{35.35} & \textbf{33.71} & \textbf{23.87}
\tabularnewline
\hline 
\end{tabular}
\end{minipage}
\vspace{-10pt}
\label{table3}
\end{table}

\begin{table}
\small
\noindent\begin{minipage}[t]{1\columnwidth}%
\tabcolsep=0.3cm
\centering
\caption{Effect of number of experts on ViSMoE.}
\begin{tabular}{c|cccccc}
\hline
\#Experts & TL\textdownarrow{} & OSR\textuparrow{} & SR\textuparrow{} & SPL\textuparrow{} & RGS\textuparrow{} & RGSPL\textuparrow{}\tabularnewline
\hline 
2 & 23.74 & 52.46 & 48.62 & 32.91 & 32.38 & 21.98
\tabularnewline
4 & 23.85 & 54.98 & 50.18 & \textbf{35.35} & 33.71 & \textbf{23.87}
 \tabularnewline
8 & \textbf{23.54} & 53.62 & 49.59 & 33.14 & 33.84 & 23.32
\tabularnewline
16 & 26.63 & \textbf{55.78} & \textbf{50.38} & 32.76 & \textbf{33.91} & 22.19
\tabularnewline
\hline 
\end{tabular}
\end{minipage}
\label{table4}
\vspace{-3mm}
\end{table}

\subsection{Ablation Studies}
We conduct the ablation studies on the val unseen split of the REVERIE dataset.\\
\textbf{Components of ViSMoE}. As shown in Table 3, we begin by evaluating the necessity of each component in ViSMoE, with the number of experts set to 4 and the top-k strategy using k=2. ``Baseline'' stands for the method of vanilla DUET \cite{chen2022think}. ``SMoE'' denotes using the sparse Mixture-of-Experts. ``Vis.'' denotes using visual-aware sparse Mixture-of-Experts. ``Aux.'' denotes using load-balance loss. Row 1 in Table 3 presents the results of the original DUET model \cite{chen2022think}. In comparison, Row 2 demonstrates improvements in all metrics except for RGSPL. Notably, when the auxiliary loss is introduced to balance the router load, all metrics show improvement over the baseline, which demonstrates the effectiveness of the vanilla SMoE and the auxiliary loss. Furthermore, Row 4 shows a substantial increase compared to Row 2, indicating that our proposed visual-aware routing policy enhances the router's ability to differentiate between various visual inputs, leading to better performance.\\
\textbf{Effect of Number of Experts on ViSMoE.} We investigate the impact of the number of experts on ViSMoE, using the top-k strategy with k=2. Table 4 illustrates the variation in results across the metrics. It can be observed that as number of experts increases beyond 4, the model size grows, but the overall performance remains nearly unchanged. This suggests that, considering the size of the dataset and the complexity of the task, having the right number of experts is more effective than simply increasing the number.\\
\textbf{Effect of Top-k on ViSMoE.} As is shown in Table 5, We study the impact of the top-k routing strategy on ViSMoE, keeping the number of experts fixed at 4. When the top-k value is set to 2 or 4, the model maintains consistent performance across different configurations. However, as top-k increases, more experts of the model are activated, leading to higher computational overhead.
\begin{table}
\small
\noindent\begin{minipage}[t]{1\columnwidth}%
\tabcolsep=0.34cm
\centering
\caption{Effect of top-k on ViSMoE.}
\begin{tabular}{c|cccccc}
\hline
top-k & TL\textdownarrow{} & OSR\textuparrow{} & SR\textuparrow{} & SPL\textuparrow{} & RGS\textuparrow{} & RGSPL\textuparrow{}\tabularnewline
\hline 
1 & 25.53 & 51.31 & 47.96 & 31.12 & 32.11 & 20.84
\tabularnewline
2 & 23.85 & \textbf{54.98} & \textbf{50.18} & \textbf{35.35} & \textbf{33.71} & 23.87
\tabularnewline
4 & \textbf{23.47} & 54.76 & 49.64 & 34.74 & 33.68 & \textbf{23.94}
 \tabularnewline
\hline 
\end{tabular}
\end{minipage}
\vspace{1pt}
\label{table5}
\end{table}
\begin{table}
\small
\noindent\begin{minipage}[t]{1\columnwidth}%
\tabcolsep=0.36cm
\centering
\caption{Effect of target layer selection.}
\begin{tabular}{c|ccccc}
\hline
Target layer & OSR\textuparrow{} & SR\textuparrow{} & SPL\textuparrow{} & RGS\textuparrow{} & RGSPL\textuparrow{}\tabularnewline
\hline 
Baseline & 51.07 & 46.98 & 33.73 & 32.15 & 23.03
\tabularnewline
Pano.1 & 51.14 & 47.16 & 32.16 & 32.04 & 22.37
\tabularnewline
Pano.2 & 53.02 & 48.63 & 32.68 & 32.23 & 22.18
 \tabularnewline
Pano.2, Cros.1 & 52.09 & 48.19 & 33.87 & 32.75 & 22.87
\tabularnewline
Pano.2, Cros.2 & 53.89 & 49.76 & 34.61 & 33.08 & 23.16
\tabularnewline
Pano.2, Cros.3 & 54.03 & \textbf{50.37} & 35.14 & 32.94 & 23.38
\tabularnewline
Pano.2, Cros.4 & \textbf{54.98} & 50.18 & \textbf{35.35} & \textbf{33.71} & \textbf{23.87}
\tabularnewline
\hline
\end{tabular}
\end{minipage}
\vspace{1pt}
\label{table6}
\end{table}
\\
\textbf{Effect of Target Layer Selection.} As in shown in Table 6, we progressively insert ViSMoE layer, with 4 experts and the top-k strategy using k=2, starting from the panorama encoder and moving to the local scale cross-modal encoder. ``Pano'' refers to the panorama encoder, while ``Cros'' indicates the local scale cross-modal encoder, where views and objects perform cross-attention over language. ``.x'' specifies the insertion of the MoE layer within the transformer blocks, from the $1^{st}$ to the $x^{th}$ block. In ``Cros'', we initialize each layer's experts' parameters using the corresponding layer's FFN parameters from LXMERT. We observe that as more and more transformer blocks in the panorama encoder and the local scale cross-modal encoder are equipped with ViSMoE layers, the model's performance gradually increases.\\

\section{Conclusion}

We introduce ViSMoE, a framework that leverages the sparse Mixture-of-Experts with a visual-aware routing policy to process diverse visual inputs. Our contributions underline the advantages of using MoE for processing visual features in embodied AI, paving the way for more robust and precise navigation systems in real-world scenarios.

\section{Acknowledgement}
This research is supported by National Natural Science Foundation of China (No.62476127), Natural Science Foundation of Jiangsu (No.BK20242039), Basic Research Program of the Bureau of Science and Technology (ILF24001), Fundamental Research Funds for the Central Universities (No.NJ2023032), Scientific Research Starting Foundation of Nanjing University of Aeronautics and Astronautics (No.YQR21022), and the High Performance Computing Platform of Nanjing University of Aeronautics and Astronautics.

%
%
%
\bibliographystyle{splncs04}
\bibliography{mybibliography}

\end{document}